\documentclass[10pt,twocolumn,letterpaper]{article}

\usepackage{iccv}
\usepackage{times}
\usepackage{epsfig}
\usepackage{graphicx}
\usepackage{amsmath}
\usepackage{amssymb}
\usepackage{bbm}
\usepackage{xcolor}
\usepackage{nth}
\usepackage{dblfloatfix}
\usepackage{algorithm}
\usepackage[noend]{algpseudocode}
\usepackage{multirow}
\newcommand{\specialcell}[2][c]{%
  \begin{tabular}[#1]{@{}c@{}}#2\end{tabular}}

\newcommand{\R}[1]{\mathbb{R}}
\newcommand{\T}[1]{\mathbb{T}}

\usepackage[pagebackref=true,breaklinks=true,letterpaper=true,colorlinks,bookmarks=false]{hyperref}

\iccvfinalcopy 

\ificcvfinal\fi
\begin{document}

\title{3D-PRNN: Generating Shape Primitives with Recurrent Neural Networks}

\author{Chuhang Zou$^\dagger$\\
\and
Ersin Yumer$^\ddagger$\\
\and
Jimei Yang$^\ddagger$\\
\and
Duygu Ceylan$^\ddagger$\\
\and
Derek Hoiem$^\dagger$
\and
$^\dagger$University of Illinois at Urbana-Champaign\\
{\tt\small \{czou4, dhoiem\}@illinois.edu}
\and
$^\ddagger$Adobe Research\\
{\tt\small \{yumer, jimyang, ceylan\}@adobe.com}
}

\maketitle

\begin{abstract}
The success of various applications including robotics, digital content creation, and visualization demand a structured and abstract representation of the 3D world from limited sensor data. Inspired by the nature of human perception of 3D shapes as a collection of simple parts, we explore such an abstract shape representation based on primitives. Given a single depth image of an object, we present \textbf{3D-PRNN}, a generative recurrent neural network that synthesizes multiple plausible shapes composed of a set of primitives. Our generative model encodes symmetry characteristics of common man-made objects, preserves long-range structural coherence, and describes objects of varying complexity with a compact representation. We also propose a method based on Gaussian Fields to generate a large scale dataset of primitive-based shape representations to train our network. We evaluate our approach on a wide range of examples and show that it outperforms nearest-neighbor based shape retrieval methods and is on-par with voxel-based generative models while using a significantly reduced parameter space.

\end{abstract}


\section{Introduction}
Many robotics and graphics applications require 3D interpretations of sensory data.  For example, picking up a cup, moving a chair, predicting whether a stack of blocks will fall, or looking for keys on a messy desk all rely on at least a vague idea of object position, shape, contact and connectedness.  A major challenge is how to represent 3D object geometry in a way that (1) can be predicted from noisy or partial observations; and (2) is useful for reasoning about contact, support, extent, and so on.  Recent efforts often focus on voxelized volumetric representations (e.g.,~\cite{yan2016perspective,wu20153d,Girdhar16b, choy20163d}).  Instead, we propose to represent objects with 3D primitives (oriented 3D rectangles, i.e. cuboids).  Compared to voxels, the primitives are much more \textit{compact}, for example 45-D for 5 primitives parameterized by scale-rotation-translation vs 32,256-D for a 32x32x32 voxel grid.  Also, primitives are \textit{holistic} --- representing an object with a few parts greatly simplifies reasoning about stability, connectedness, and other important properties.  Primitive-based 3D object representations have long been popular in psychology (e.g. ``geons'' by Biederman~\cite{biederman1987recognition}) and interactive graphics (e.g. ``Teddy''~\cite{igarash1999teddy}), but they are less commonly employed in modern computer vision due to the challenges of learning and predicting models that consist of an arbitrary number of parameterized components. 

\begin{figure}
\begin{center}
\includegraphics[width=0.9\linewidth]{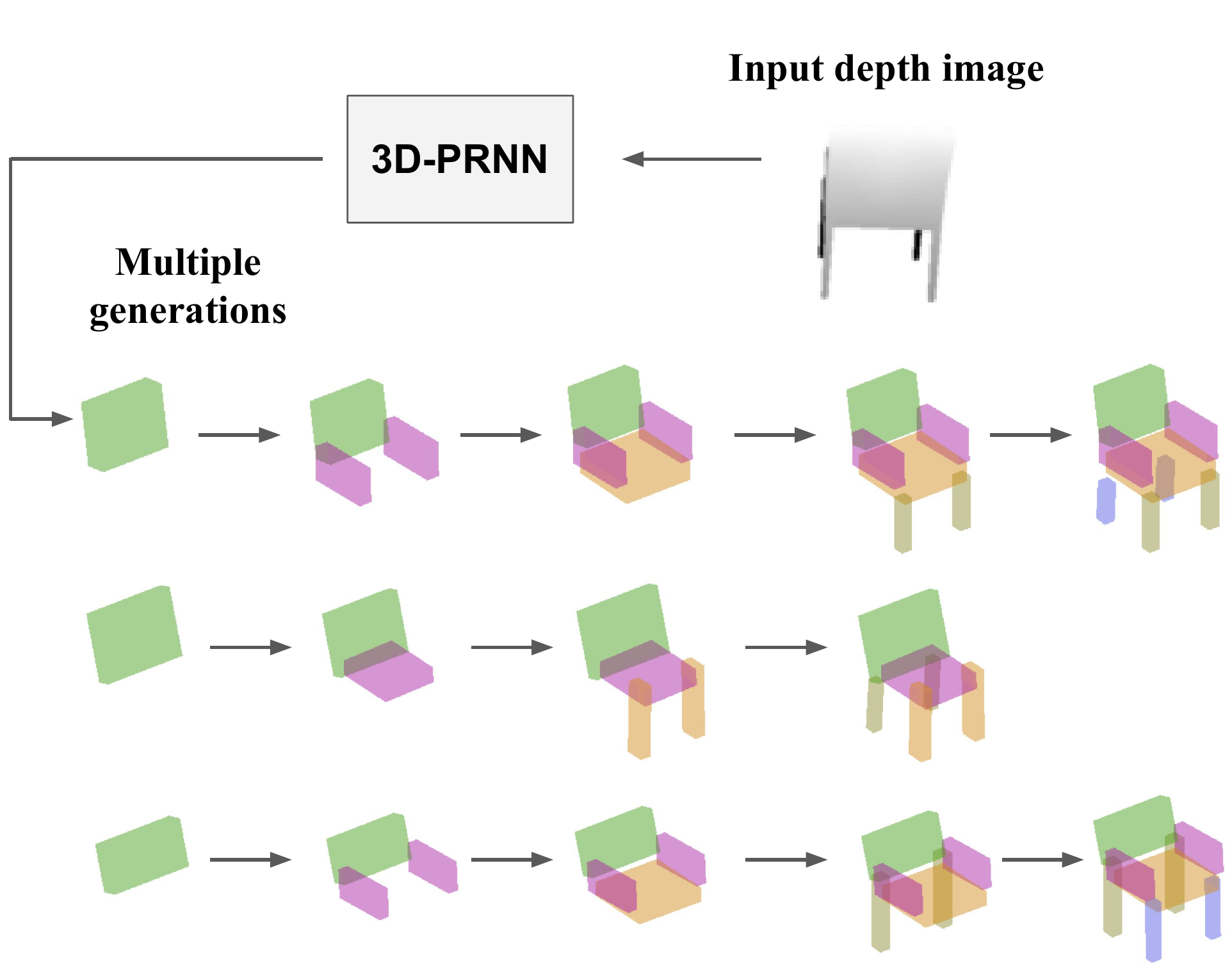}
\end{center}
\vspace{-0.15in}
   \caption{\label{fig:intro} A step-by-step primitive-based shape generation by 3D-PRNN. As an illustration, given single depth image, we sequentially predicts sets of primitives that form the shape. Each time we randomly sample one primitive from a set and generate the next set of primitives conditioning on the current sample.
   }
\vspace{-0.15in}
\end{figure}

 \begin{figure*}
\begin{center}
\includegraphics[width=0.97\linewidth]{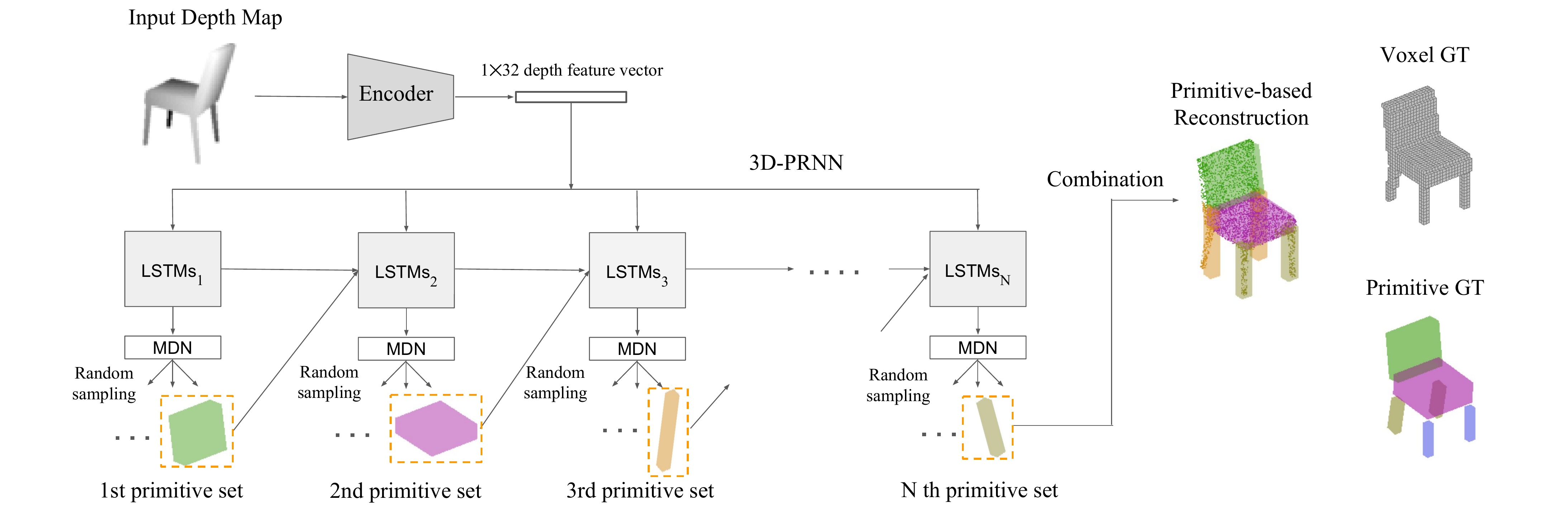}
\end{center}
\vspace{-0.05in}
   \caption{\label{fig:overview} \textbf{3D-PRNN overview}. We illustrate the method on the task of single depth shape completion. The network starts from encoding the input depth image into a feature vector, which is then sent to the "recurrent generator" consisting stacks of Long Short-Term Memory (LSTM) and a Mixture Density Network (MDN). At each time step, the network predicts a set of primitives conditioned on both the depth feature and the previously sampled single primitive. The final reconstruction result and ground truth are shown on the right.}
   \vspace{-0.1in}
\end{figure*}

Our goal is to learn 3D primitive representations of objects from unannotated 3D meshes.  We follow an encoder-decoder strategy, inspired by recent work~\cite{DBLP:journals/corr/Graves13,DBLP:conf/icml/OordKK16}, using a recursive neural network (RNN) to encode an implicit shape representation and then sequentially generate primitives to approximate the shape as shown in Fig.~\ref{fig:intro}.  
One challenge in training such a primitive generation network is acquiring ground truth data for primitive-based shape representations. To address this challenge, we propose an efficient method based on Gaussian Fields and energy minimization~\cite{boughorbel2010new} to iteratively parse shapes into primitive components. We optimize a differentiable loss function using robust techniques (L-BFGS\cite{zhu1997algorithm}). We use this (unsupervised) optimization process to create the primitive ground truth, solving for a set of primitives that approximates each 3D mesh in a collection.  The RNN can then be trained to generate new primitive-based shapes that are representative of an object class' distribution or to complete an object's shape given a partial observation such as a depth image or point cloud.  

To model shapes, we propose 3D-PRNN, an RNN-based generative model that predicts context-sensitive sequences of primitives in object-centric coordinates, as shown in Figure~\ref{fig:overview}.  
To predict shape from depth, the network is trained jointly with the single depth image and a sequence of primitives configurations (shape, translation and rotation) that form the complete shape. During testing, the network gets input of a depth map and sequentially predicts primitives (ending with a stop signal) to reconstruct the shape. Our generative RNN architecture is based on a Long Short-Term Memory (LSTM) and a Mixture Density Network (MDN).

We evaluate our proposed generative model by comparing with baselines and the state-of-the-art methods. We show that, even though our method has less degrees of freedom in representation, it achieves comparable accuracy with voxel based reconstruction methods. We also show that encoding further symmetry and rotation axis constraints in our network significantly boosts performance. 

Our main contributions are:
\begin{itemize}
\item We propose 3D-PRNN: a generative recurrent neural network that reconstructs 3D shapes as sequences of primitives given a single depth image.
\item We propose an efficient method to fit primitives from point clouds based on Gaussian-fields and energy minimization. Our primitive representation provides enough training samples for 3D-PRNN in 3D reconstruction.
\end{itemize}



\section{Related Work}
\textbf{Primitive-based Shape Modeling}:  Biederman, in the early 1980s, popularized the idea of representing shapes as a collection of components or primitives called ``geons''~\cite{biederman1987recognition}, and early computer vision algorithms attempted to recover object-centered 3D volumetric primitives from single images~\cite{dickinson1992primitive}. In computer aided design, primitive-based shape representations are used for 3D scene sketches~\cite{zeleznik2007sketch, schmidt2007shapeshop}, shape completion from point clouds~\cite{schnabel2009completion,li2011globfit,schnabel2010efficient}. In the case that scans of shapes often have canonical parts like planes or boxes and efficient solution for large data is required, primitives are used in reconstructions of urban and architectural scenes~\cite{chauve2010robust, lafarge2013surface, bodis2014fast,sinha2008interactive}. 
Recently, more compact and parametric representations in the form of template objects~\cite{wu2016single}, and set of primitives~\cite{tulsiani2016learning} have been introduced. These representations, however, require non trivial effort to accommodate variable number of configurations within the object class they are trained for. This is mainly because of their single feed-forward design, which implicitly forces the prediction of a discrete number of variables at the same time. 

 \begin{figure}
\begin{center}
\includegraphics[width=0.8\linewidth]{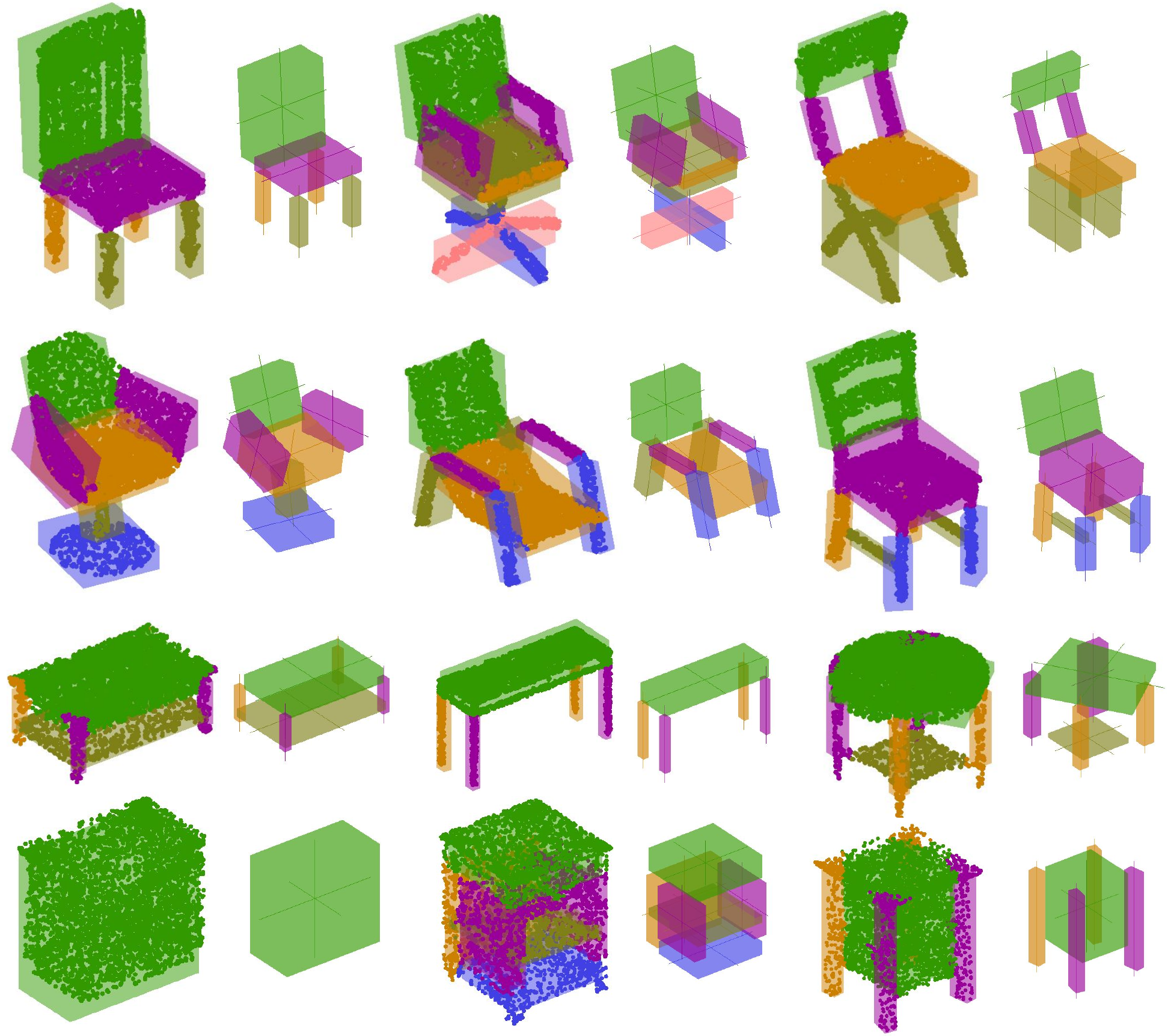}
\end{center}
\vspace{-0.05in}
   \caption{\label{fig:fitting} \textbf{Sample primitive fitting result}. We show our primitive fitting results on chairs, tables and sofas. We overlay our fitted primitives on the sampled 3D point clouds of each shape.}
\vspace{-0.15in}
\end{figure}

 \begin{figure}
\begin{center}
\includegraphics[width=0.76\linewidth]{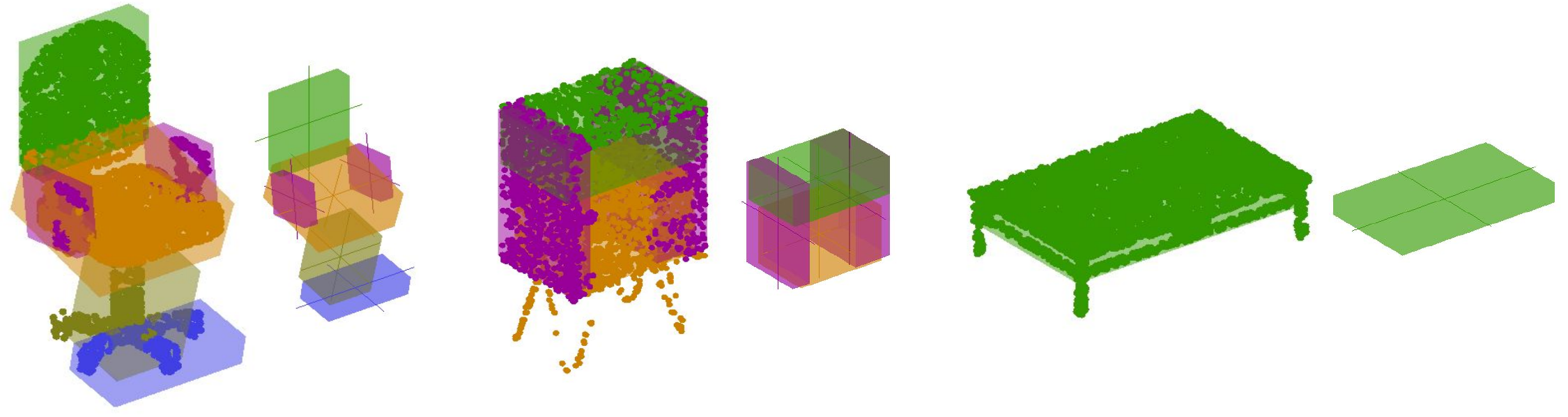}
\vspace{-0.05in}
\end{center}
   \caption{\label{fig:failure} \textbf{Failure cases}. Main causes are : too complex shape details to be represented by primitive blocks (left),  The smoothing property of Gaussian force fields is not good at describing small hollow shape (middle), small cluster of point clouds are easily missed through our randomized search scheme (middel and right). }
\vspace{-0.15in}
\end{figure}
 
\textbf{Object 3D shape reconstruction} can be attempted given an RGB image ~\cite{yan2016perspective,Girdhar16b, aubry2014seeing,choy20163d}or depth image~\cite{wu20153d, rock2015completing, firman2016structured} Recently proposed representations and prediction approaches for 3D data in the context of prediction from sensory input have mainly either focused on part- and object-based retrieval from large repositories~\cite{rock2015completing,aubry2014seeing, lim2013parsing, kholgade20143d}, or voxelized volumetric representations~\cite{yan2016perspective,wu20153d,Girdhar16b, choy20163d}. A better model fitting includes part deformation~\cite{chen20133} and symmetry~\cite{kurz2014symmetry}. Wu et al.~\cite{wu20153d} present preliminary results on automatic shape completion from depth by classifying hidden voxels with a deep network. Wu et al.~\cite{wu2016single} reconstruct shapes based on predicted skeletons. Unlike mesh-based or voxel-based shape reconstruction, our method predicts shapes with aggregations of primitives that has the benefit for lower computational and storage cost.

\textbf{Generative Models with RNNs: } Graves~\cite{DBLP:journals/corr/Graves13} uses Long Short-term Memory recurrent neural networks to generate complex sequences of text and online handwriting. Gregor et al.~\cite{DBLP:conf/icml/GregorDGRW15} combine LSTM and a variational auto-encoder, called the Deep Recurrent Attentive Writer (DRAW) architecture, for image generation. The DRAW architecture is a pair of RNNs with an encoder network that compresses the real images presented during training, and a decoder that reconstitutes images after receiving codes. Rezende et al.~\cite{NIPS2016_6600} extend DRAW to learn generative models of 3D structures and recover this structure from 2D images via probabilistic inference. Our 3D-PRNN, which sequentially generates primitives, is inspired by Graves' work to sequentially generate parameterized handwriting strokes and the PixelRNN approach~\cite{DBLP:conf/icml/OordKK16} to model natural images as sequentially generated pixels.  To produce parameterized 3D primitives (oriented cuboids), we customize the RNN to encode explicit geometric constraints of symmetry and rotation.  For example, separately predicting whether a primitive should rotate along each axis and by how much improves results over more simply predicting rotation values, since many objects consist of several (unrotated) cuboids. 

\section{Fitting Primitives from Point Clouds}\label{parsing}
One challenge in training our 3D-PRNN primitive generation network is the lack of large scale ground truth primitive based shape reconstruction data. 
We propose an efficient method to 
generate such data. 
Given a point cloud representation of a shape, our approach finds the most plausible primitives to fit in a sequential manner, e.g. given a table, the algorithm might identify the primitive that fits to the top surface first and then the legs successively. We use rectangular cuboids as primitives which provide a plausible abstraction for most man-made objects. Our method proposes a fast parsing solution to decompose shapes with varying complexity into a set of such primitives. 

\subsection{Primitive Fitness Energy}
We formulate the successive fitting of primitives as an energy minimization scheme. While primitive fitting at each step resembles the method of Iterative Closest Point (ICP)~\cite{Besl:1992}, we have additional challenges. ICP ensures accurate registration when provided with a good initialization, but in our case we have no prior knowledge about the number and the rough shape of the primitives. Moreover, we need to solve the more challenging partial matching problem since each primitive matches only part of the shape, which we do not know in advance. 

We represent the shape of each primitive with scale parameters $S=[s_x, s_y, s_z]$, which denotes the scale of a unit cube along three orthogonal axes. The position and orientation of the primitive are represented by translation, $T=[t_x,t_y,t_z]$, and Euler angles, $\theta=[\theta_x, \theta_y, \theta_z]$, respectively. Thus the primitive is parameterized by $x=[s_x, s_y, s_z, t_x, t_y, t_z, \theta_x, \theta_y, \theta_z]$. Furthermore, we assume a fixed sampling of the unit cube into a set of points, $P = \{p_m\}_{m = 1,\ldots,M}$. Given a point cloud representation of a shape, $Q = \{q_n\}_{n = 1,\ldots,N}$, our goal is to find the set of primitives $X=\{x_t\}_{t=1,2,3,...}$ that best fit the shape. We employ the idea of Gaussian Force Fields~\cite{boughorbel2010new} and Truncated Signed Distance Function (TSDF)~\cite{newcombe2011kinectfusion} to formulate the following continuously differentiable energy function which is convex in a large neighborhood of the parameters:

\begin{align}\label{cost}
E_p = -\sum_{m,n}V_p\min\bigg(\exp\Big(-\frac{\|R(\theta) S p_m + T - q_n\|^2}{\sigma^2}\Big), \xi \bigg),
\end{align}
where $R(\theta)$ is the rotation matrix, $\xi$ is the truncation parameter ($\xi = 0.9$ in our experiments) and $V_p$ denotes the volumetric-wise sampling ratio that is calculated as the volume of primitive $P$ over its number of sampled points $M$. $V_p$ helps avoid local minimum that results in a too small or too large primitive. Our formulation represents the error as a smooth sum of Gaussian kernels, where far away point pairs are penalized less to account for partial matching.

The energy function given in Eq.~\ref{cost} is sensitive to the parameter $\sigma$. A larger $\sigma$ will encourage fitting of large primitives while allowing larger distances between matched point pairs. In order to prefer tighter fitting primitives, we introduce the concept of \emph{negative shape}, $Q^-$, which is represented as a set of points sampled in the non-occupied space inside the bounding box of a shape. We update our energy function as: 
\begin{align}\label{n_cost}
E_w = E_P^+ - \alpha E_P^-,
\end{align}
where $E_P^+$ is the fitting energy between the shape and the primitive and $E_P^-$ is the fitting energy between the negative shape and the primitive. Given point samples, both $E_P^+$ and $E_P^-$ are computed as in Eq.~\ref{cost}. $\alpha$ denotes the relative weighting of these two terms and is defined as $\alpha = \max\big(\min(10, |Q|/|Q^-|\times 5),0.1\big)$.
 
\subsection{Optimization}
Given the energy formulation described in the previous section, we perform primitive fitting in a sequential manner. During each iteration, we randomly initialize $10$ primitives, optimize Eq.~\ref{n_cost} for each of these primitives and add the best fitting primitive to our primitive collection. We then remove the points in $Q$ that are fit by the selected primitive and iterate. We stop once all the points in $Q$ are fit by a primitive. We optimize Eq.~\ref{n_cost} in an iterative manner. We first fix $\theta$ and solve for $S$ and $T$, we then fix $S$ and $T$ and solve for $\theta$. In our experiments this optimization converges in $5$ iterations and we use the L-BFGS toolbox~\cite{schmidt2012minfunc} at each optimization step. We summarize this process with the pseudo-code given in Alg.~\ref{euclid}. 

\begin{algorithm}
\caption{Primitive fitting}\label{euclid}
\begin{algorithmic}[1]
\item Given shape point clouds Q and empty primitive set X;
\item $ \beta= 0.97 |Q|$, t = 0;
\While{$|Q|<\beta$ or $i < $maxPrimNum}
\State{$E_{best} =$ Inf};
\For{$i = 1:$maxRandNum}
\State{$\theta=[0,0,0]$, random initialize $S$, $T$, $j = 0$};

\While{$\delta <0.01 $ or $j < $maxIter}
\State{fix $\theta$, solve $S$, $T\rightarrow S^*$, $T^*$ by Eq~.\ref{n_cost}};
\State{fix $S^*$, $T^*$, update $\theta\rightarrow \theta^*$ by Eq~.\ref{n_cost}};
\State{calculate $E_w(S^*, T^*, \theta^*)$ by Eq~.\ref{n_cost}};
\If{$E_w < E_{best}$}
\State{$E_{best} = E_w$, $x_{best} = [S^*, T^*, \theta^*]$};
\EndIf
\State{$\delta = \|[S, T, \theta] - [S^*, T^*, \theta^*]\|^2$};
\State{$S = S^*$, $T = T_p^*$, $k = k+1$};
\EndWhile
\EndFor
\State{$x_t = x_{best}$}, add $x_t$ to $X$, $t = t+1$;
\State{Remove fitted points from $Q$ and add to non-occupied space $Q^-$}
\EndWhile
\Return{$X$}
\end{algorithmic}
\end{algorithm}

\textbf{Simplification with symmetry.} We utilize the symmetry characteristics of man-made shapes to further speed up the primitive parsing procedure. We use axis-aligned 3D objects where symmetric objects have a common global symmetry plane. We compare the geometry on the two sides of this plane to decide whether an object is symmetric or not. Once we obtain a primitive that lies on one side of the symmetry plane, we automatically generate the symmetric primitive on the other side of the plane.

\textbf{Refinement.} At each step, we fit primitives with a relatively larger Gaussian field~($\sigma = 2$ in Eq.~\ref{cost}) for fast convergence and easier optimization. We then refine the fitting with a finer energy space~($\sigma = 0.5$) to match the primitive to the detailed shape of the object. While our random search scheme enables a fast parsing method, errors may accumulate in the final set of primitives. To avoid such problems, we perform a post-refinement step. We refine the parameters of a single primitive $x_t$ while fixing the other parameters. We use the parameters of $x_t$ obtained from the initial fitting as initialization. We define the energy terms in Eq.~\ref{n_cost} with respect to the points that are fit by $x_t$ and the points that are not fit by any primitive yet. We note that this sequential refinement is similar to back propagation used to train neural networks. In our experiments, we perform the refinement each time we fit 3 new primitives.

\section{3D-PRNN: 3D Primitive Recurrent Neural Networks}
Generating primitive-based 3D shapes is a challenging task due to the complex multi-modal distribution of shapes and the unconstrained number of primitives required to model such complex shapes. We propose 3D-PRNN, a generative recurrent neural network to accomplish this task. 3D-PRNN can be trained to generate novel shapes both randomly and by conditioning on partial shape observations such as a single depth map. 


 \begin{figure}
\begin{center}
\includegraphics[width=0.95\linewidth]{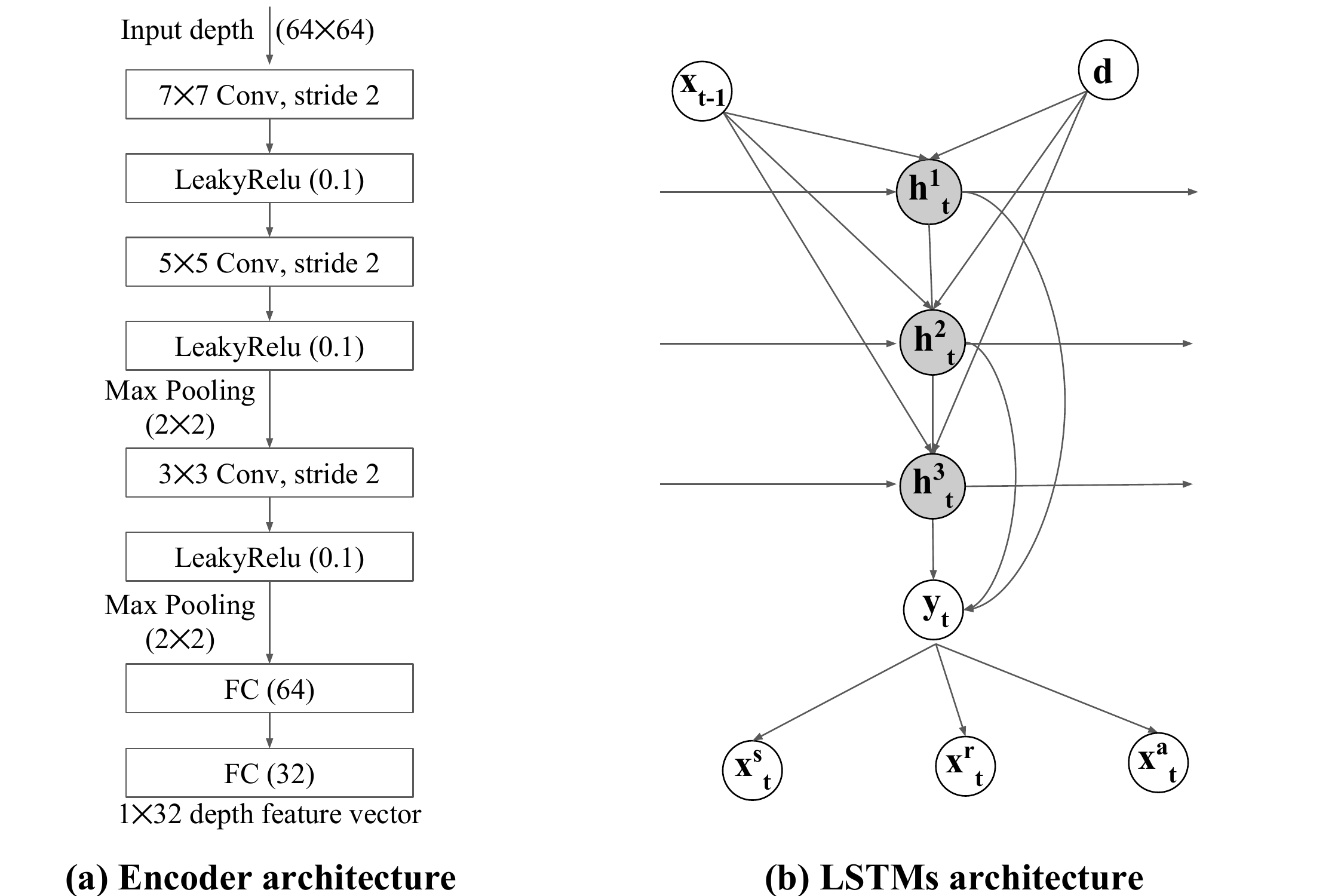}
\end{center}
\vspace{-0.05in}
   \caption{\label{fig:lstm} Detailed architectures of (a) the depth map encoder and (b) the primitive recurrent generator unit in \textbf{3D-PRNN}. See the architecture descriptions in Section 4.1.}
\vspace{-0.15in}
\end{figure}

\subsection{Network Architecture}
An overview of the 3D-PRNN network is illustrated in Fig.~\ref{fig:overview}. The network gets as input a single depth image and sequentially predicts primitives to form a 3D shape. For each primitive, the network predicts its shape (height, length, width), position (i.e. translation), and orientation (i.e. rotation). Additionally, at each step, a binary \emph{end of generation} signal is predicted which indicates no more primitive should be generated. 

\textbf{Depth map encoder.} Each input depth map, $I$, is first resized to be $64\times 64$ in dimension with values in the range $[0,1]$~(we set the value of background regions to 0). $I$ is passed to an encoder which consists of stacks of convolutional and LeakyRelu~\cite{maas2013rectifier} layers as shown in Fig.~\ref{fig:lstm}~(a): the first layer has $32$ kernels of size $7\times 7$ and stride $2$, with a LeakyRelu layer of $0.1$ slope in the negative part. The second layer consists of $64$ kernels of size $5\times 5$ (stride $2$), followed by the same setting of LeakyRelu and a max pooling layer. The third layer has $128$ kernels of size $3\times 3$ (stride $2$) followed by LeakyRelu and max pooling. The next two fully-connected layers has neurons of $64$ and $32$. The output $1\times 32$ feature vector $d$ is then sent to the recurrent generator to predict a sequence of primitives.

\textbf{Recurrent generator.} We apply the Long Short-Term Memory (LSTM) unit inside the recurrent generator, which is shown to be better at alleviating the vanishing or exploding gradient problems~\cite{pascanu2013difficulty} when training RNNs. The architectural design is shown in Fig.~\ref{fig:lstm}~(b). The prediction unit consists of $L$ layers of recurrently connected hidden layers~(we set $L = 3$, which is found to be sufficient to model the complex primitive distributions) that encode both the depth feature $d$ and the previously predicted primitive $x_{t-1}$ and then computes the output vector, $y_t$. $y_t$ is used to parametrize a predictive distribution $Pr(x_t|y_t)$ over the next possible primitive $x_t$. The hidden layer activations are computed by iterating over the following equations in the range $t = [1,T]$ and $l = [2,L]$:
\begin{align}
z_{t}^l &= W_{x}^l x_{t-1} + W_{h}^l h_{t-1}^l + W_{c}^l h_{t}^{l-1} + W_{d}^l d\\
[i_t^l, f_t^l, o_t^l] &= \sigma(z_{t}^l)\\
g_t^l &= \tanh(z_{t}^l)\\
c_t^l &= f_t^lc_{t-1}^l + i_t^lg_t^l\\
h_t^l &=o_t^l\tanh(c_t^l)
\end{align}
where $z_{t}^l$ capsules the input features in the $l$-th layer (when $l=1$, there is no hidden value propagated from the previous layers and thus $z_{t}^1 = W_{x}^1 x_t + W_{h}^1 h_{t-1}^1+ W_{c}^1d$), $h_t$ and $c_t$ denote the hidden and cell states, whereas $W_{x}^l, W_{h}^l, W_{c}^l, W_{d}^l$ denote the linear weight matrix (we omit the bias term for brevity), $i_t$, $f_t$, $o_t$, $g_t$ are respectively the input, forget, output, and context gates, which have the same dimension as the hidden states (size of $400$). $\sigma$ is the logistic sigmoid function and $\tanh$ is the hyperbolic tangent function. 

At each time step $t$, the distribution of the next primitive is predicted as $y_t = W_{y}[h_t^1, h_t^2, ..., h_t^L]$, where we perform a linear transformation on the concatenation of all the hidden values. This concatenation is similar in spirit to using \emph{skip connections}~\cite{srivastava2015training,he2016deep}, which is shown to help training and mitigate the ‘vanishing gradient’ problem. In a similar fashion, we also pass the depth feature $d$ to all the hidden layers. We will explain latter how the primitive configuration $x_t$ is sampled from a distribution predicted from $y_t$.

We predict parameters of one axis per time conditioned on the previous axis. We model this joint distribution of parameter on each axis $x^s_i=[s_i,t_i]$ (where $i$ indicates one of the 3 axes of space) as a mixture of Gaussians conditioned on previous axis
with $K$ mixture components:
\begin{align}
\big(\{\pi_t^k, \mu_t^k, \sigma_t^k, \rho_t^k\}^K_{k = 1}, e_t\big) = f(y_t),
\end{align}
where $\pi_t$, $\mu_t$, $\sigma_t$ and $\rho_t$ are the weight, mean, standard deviation, and correlation of each mixture component respectively, predicted from a fully connected layer $f(y_t)$. Note that $e_i$ is the binary stopping sign indicating whether the current primitive is the final one and it helps with predicting a variable-length sequence of primitives. In our experiments we set $K = 20$. We randomly sample a single instance $x^s_i = [s_i, t_i, e_i]\in \mathbb{R}\times \mathbb{R}\times \{0,1\}$ drawn from the distribution $f(y_t)$.  
The sequence $x_t$ represents the parameters in the following order: $x^s_x \rightarrow x^s_y \rightarrow x^s_z$ for the shape translation configuration on $x,y,z$ axis of the first primitive and the stopping sign.

This is essentially a mixture density network (MDN)~\cite{bishop1994mixture} on top of the LSTM output and its loss is defined:
\begin{align}\label{loss_mdn}
&L_s(x) = \sum^T_{t = 1} - \log\big(\sum_k\pi_t^k N(x_{t+1}|\mu_t^k, \sigma_t^k, \rho_t^k)\big)\nonumber\\
&-\mathbb{I}\big((x_{t+1})_3 = 1\big)\log e_t -\mathbb{I}\big((x_{t+1})_3 \ne 1\big)\log (1-e_t)
\end{align}
The MDN is trained by maximizing the log likelihood of ground truth primitive parameters in each time step, where we calculate gradients explicitly for backpropagation as shown by Graves~\cite{DBLP:journals/corr/Graves13}. We found this stepwise supervised training works well and avoids sequential sampling used in \cite{tulsiani2016learning, eslami2016attend}. 

\begin{figure}
\begin{center}
\includegraphics[width=0.99\linewidth]{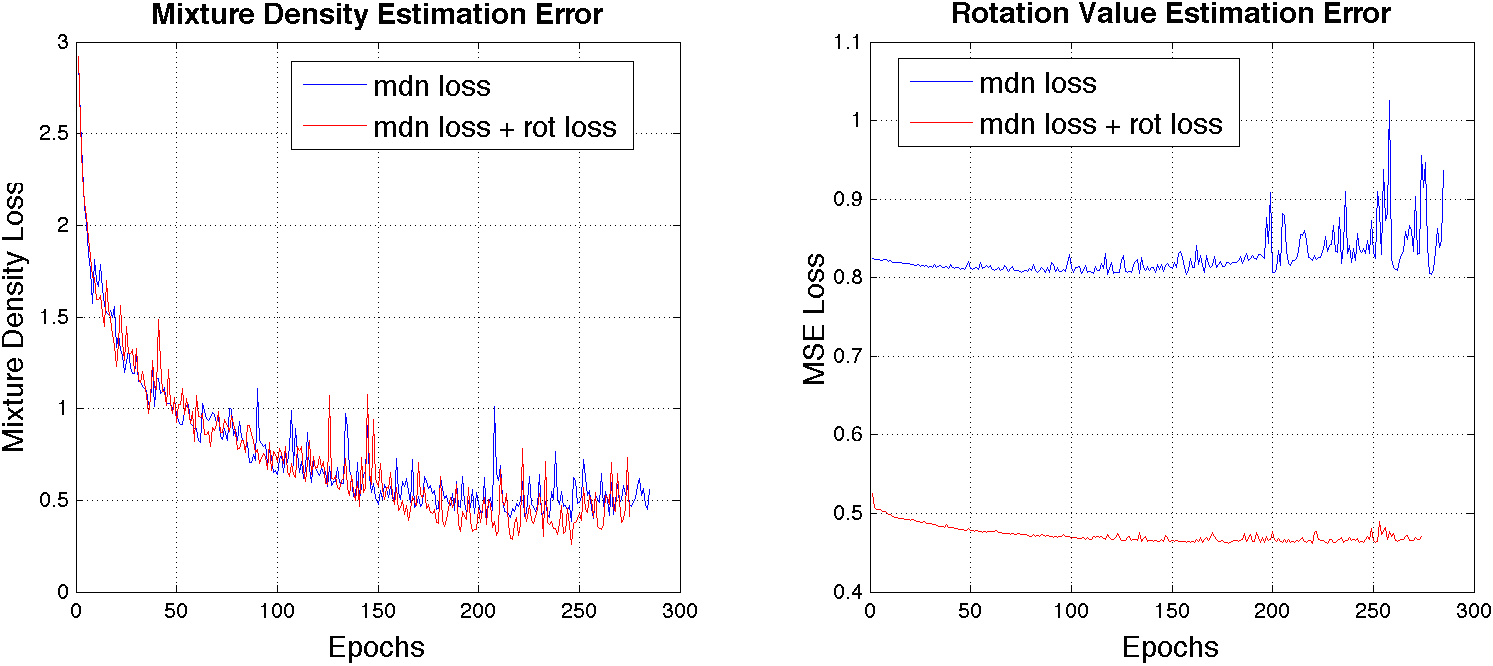}
\end{center}
\vspace{-0.05in}
   \caption{\label{fig:learning} \textbf{Training performance comparison on validation set of synthetic depth map from ModelNet.} Both the mixture density loss and the rotation MSE loss are averaged by sequence length. The rotation values are normalized and values can have ranges around 13, compared with the $<1$ MSE loss. Our mixture density estimation and rotation value estimation performs better by enforcing loss on predicting rotation axis.}
   \vspace{-0.15in}
\end{figure}

 
\textbf{Geometric constraints.} 
Another challenge in predicting primitive-based shape is to model rotation, given that the rotation axis is sensitive to slight change in rotation values under Euler angles. 
We found that by jointly predicting the rotation axis $x^a$ and the rotation value $x^r$, both the rotation prediction performs better and the overall primitive distribution modeling get alleviated as shown in Fig.~\ref{fig:learning}, quantitative experiments are in Sec.~\ref{exp:rec}. The rotation axis~($x^{a}$) is predicted by a three-layered fully connected network $g(y_t)$ with size $64$,$32$ and $1$ and sigmoid function as shown in fig.~\ref{fig:lstm}. The rotation value~($x^{r}$) is predicted by a separate three-layered fully connected network $g^*(y_t)$ with size $64$, $32$ and $1$ and a $Tanh(\cdot)$ function. 

\subsection{Loss Function}
The overall sequence loss of our network is:
\begin{align}
&L(x) = L_s(x^s) + L_r(x^r) + L_a(x^a),\\
&L_r(x) = \sum_t \|x^r_t - \hat{x^r_t}\|^2 \\
&L_a(x) = \sum_t \|x^a_t - \hat{x^a_t}\|^2
\end{align}
$L_s(x)$ is defined in Eq.~\ref{loss_mdn}. $L_r(x)$ is a mean square loss between predicted, $x^r_t$, and target, $\hat{x^r_t}$, rotation. $L_a(x)$ is the mean square loss between the predicted, $x^a_t$, and ground truth, $\hat{x^a_t}$, rotation axis.

 \begin{figure}
\begin{center}
\includegraphics[width=0.8\linewidth]{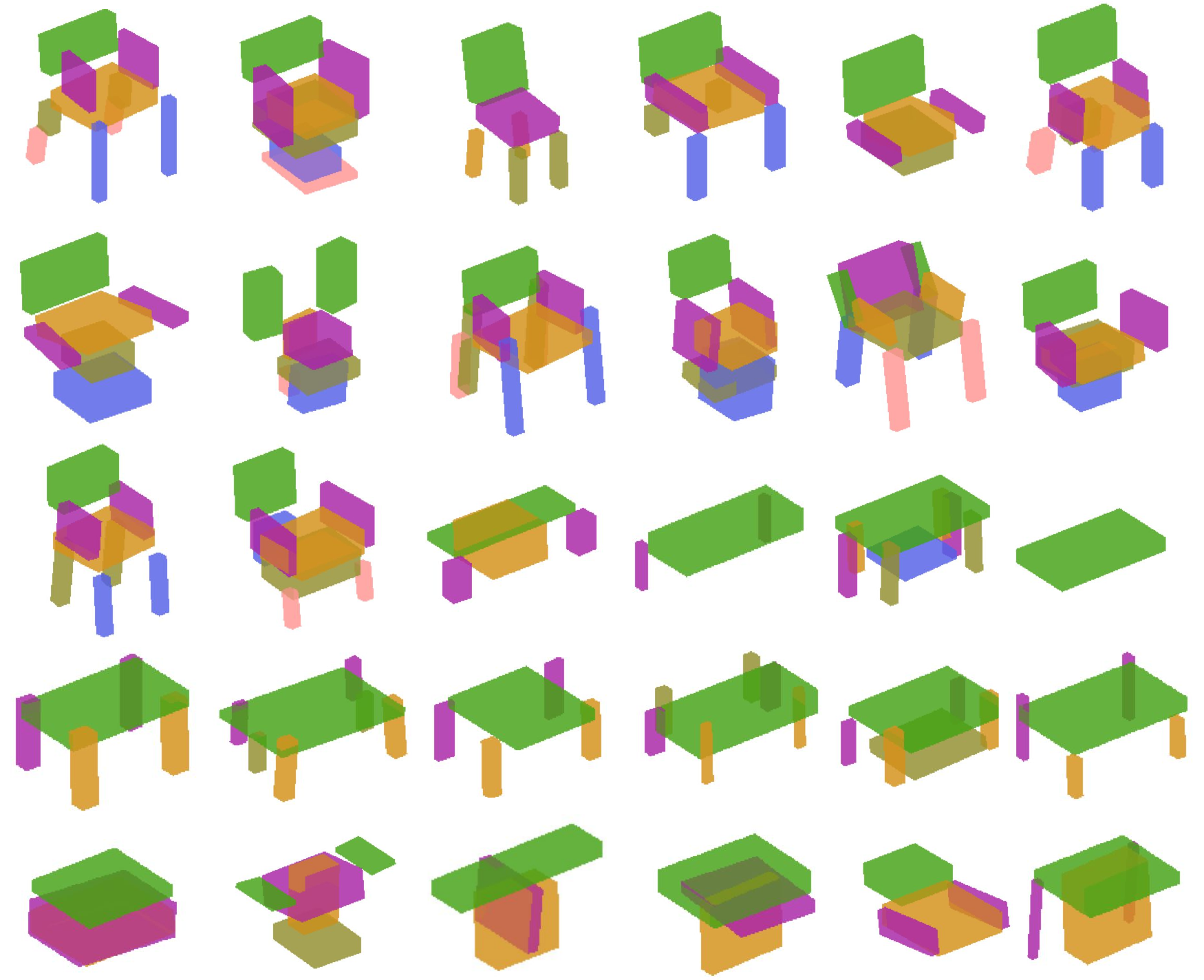}
\end{center}
\vspace{-0.05in}
   \caption{\label{qual:syn} \textbf{Shape synthesis result}. We show various random sampled shapes by our 3D-PRNN. The network is trained and tested without context input. The coloring indicates the prediction order. }
   \vspace{-0.15in}
\end{figure}

 \begin{figure*}
\begin{center}
\includegraphics[width=0.92\linewidth]{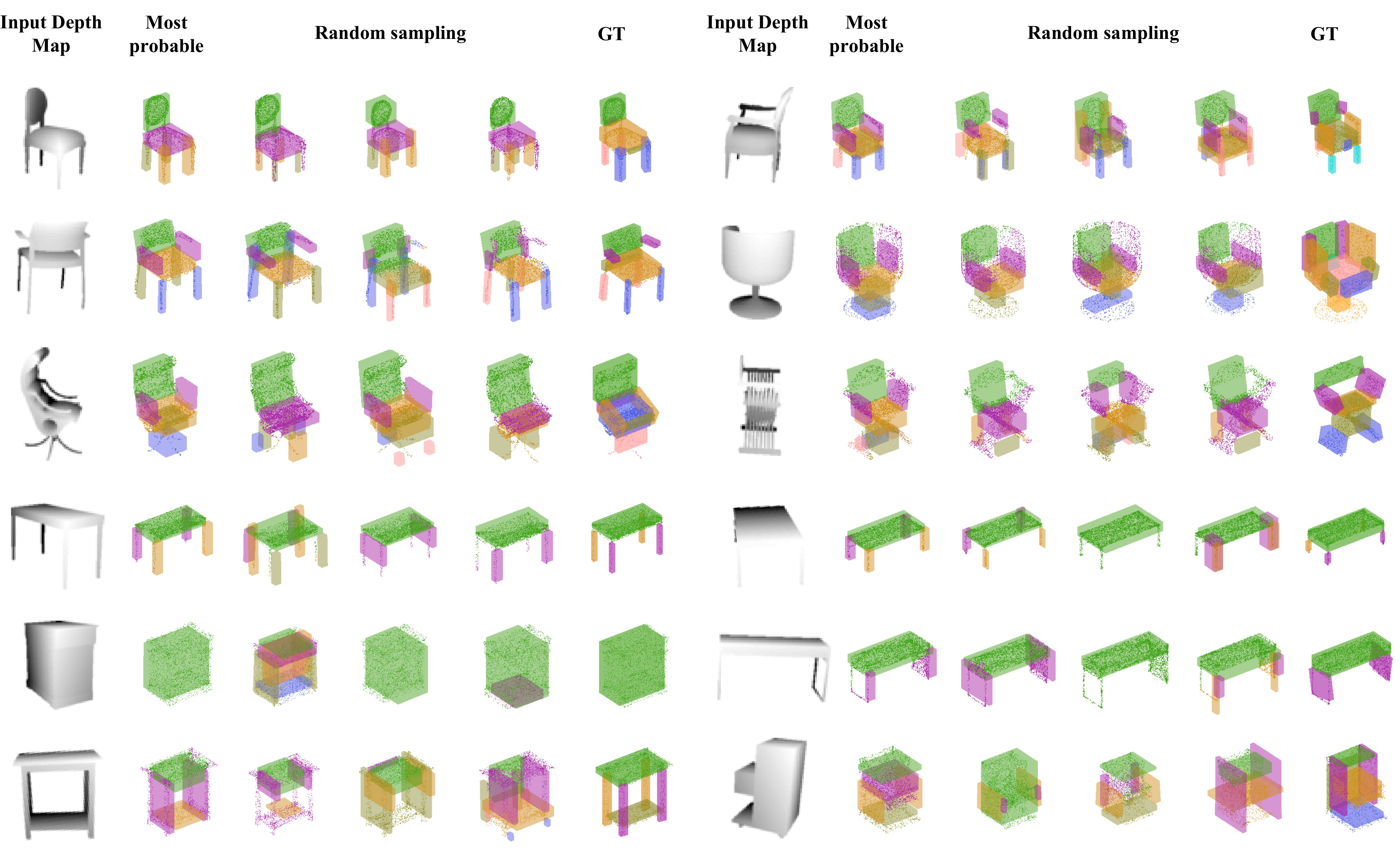}
\end{center}
\vspace{-0.15in}
   \caption{\label{qual:rec} \textbf{Sample reconstruction from synthetic data from ShapeNet}. We show the input depth map, with the most probable shape reconstruction from 3D-PRNN, and three successive random sampling results, compared with our ground truth primitive representation.}
   \vspace{-0.1in}
\end{figure*}

 \begin{figure*}
\begin{center}
\includegraphics[width=0.9\linewidth]{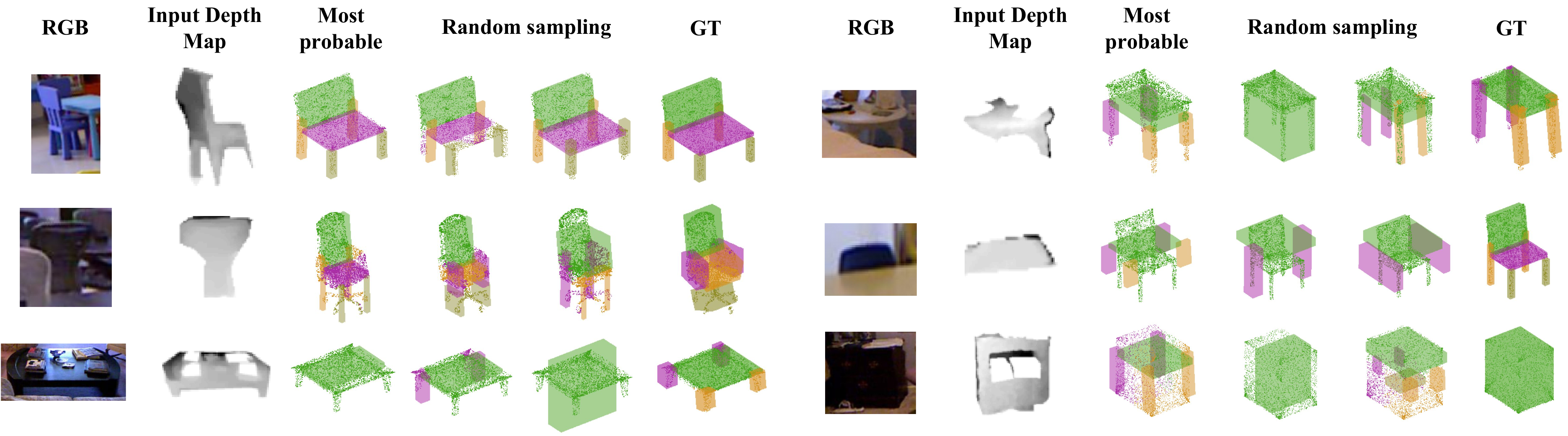}
\end{center}
\vspace{-0.15in}
   \caption{\label{qual:real} \textbf{Sample reconstruction from real depth map in NYUdv2}. We show the input depth map, with the most probable shape reconstruction from 3D-PRNN, and two successive random sampling results, compared with our ground truth primitive representation.}
   \vspace{-0.15in}
\end{figure*}

\section{Experiments and Discussions}
We show quantitative results on automatic shape synthesis. We quantitatively evaluate our 3D-PRNN in two tests: 1) 3D reconstruction on synthetic depth maps and 2) using real depth maps as input.

We train our 3D-PRNN on ModelNet~\cite{wu20153d} categories: 889 chairs, 392 tables and 200 nightstands. We employ the provided another 100 testing samples from each class for evaluation. We train a single network with all shapes classes jointly. In all experiments, to avoid overfitting, we hold out $15\%$ of the training samples, which are then used to choose the number of training epochs. We then retrain the network using the entire training set. Since a single network is trained to encode all three classes, when predicting shape from depth images, for example, there is an implicit class prediction as well. 

\subsection{Implementation}\label{training}
We implement 3D-PRNN network using Torch. We train our network on primitive-based shape configurations generated as described in Sec.~\ref{parsing}. The parameters of each primitive (i.e. shape, translation and rotation) are normalized to have zero mean and standard deviation. We observe that the order of the primitives generated by the method described in Sec.~\ref{parsing} involves too much randomness that makes training hard. Instead, we pre-sort the primitives based on the height of each shape center in a decreasing fashion. This simple sorting strategy significantly boosts the training performance. Additionally, our network is trained only on one side of the symmetric shapes to shorten the sequence length and speed up the training process. To train with the generative mechanism, we use simple random sampling technique. We use ADAM~\cite{DBLP:journals/corr/KingmaB14} to update network parameters with a learning rate of $0.001$, $\alpha = 0.95$, and $\epsilon = e^{-6}$. We train the network with batch size $380$ and $50$ on the synthetic data and on the real data respectively.

At test time, the network takes a single depth map and sequentially generates primitives until a stop sign is predicted. To initialize the first RNN feature $x$, we perform a nearest neighbor query based on the encoded feature of the depth map to retrieve the most similar shape in the training set and use the configuration on its first primitive. 

\subsection{Shape Synthesis}
3D-PRNN can be trained to generate new primitive-based shapes. Fig.~\ref{qual:syn} shows our randomly generated shapes synthesized from all three shape classes. We initialize the first RNN feature $x$ with a random sampled primitive configuration from the training set. Since the first feature corresponds to ``width'', ``translation in x-axis'', and ``rotation on x-axis'' of the primitive, formally this initialization process is defined as drawing a sample from a discrete uniform distribution of these parameters where the discrete samples are constructed from the training examples. The figure shows that 3D-PRNN can learn to generate representative samples from multiple classes and sometimes creates hybrids from multiple classes.

\subsection{Shape Reconstruction from Single Depth View}\label{exp:rec}
\textbf{Synthetic data.} We project synthetic depth maps from training meshes. For both training and testing, we perform rejection-sampling on a unit sphere for 5 views, bounded within 20 degrees of the equator.  The complete 3D shape is then predicted using a single depth map as input to 3D-PRNN. Our model can generate a sampling of complete shapes that match the input depth, as well as the most likely configuration, determined as the mean of the Gaussian from the most probable mixture.  We report 3D intersection over union (IoU) and surface-to-surface distance~\cite{rock2015completing} of the most likely predicted shape to the ground truth mesh. To compute IoU, the ground truth mesh is voxelized to 30 x 30 x 30 resolution, and IoU is calculated based on whether the voxel centers fall inside the predicted primitives or not. Surface-to-surface distance is computed using 5,000 points sampled on the primitive and ground truth surfaces, and the distance is normalized by the diameter of a sphere tightly fit to the ground truth mesh (e.g. 0.05 is $5\%$ of object maximum dimension).  

Tables~\ref{iou-synthetic} and~\ref{surf-synthetic} show our quantitative results. ``GT prim'' is the ground truth primitive representation generated by our parsing optimization method during training. This serves as an upper bound on performance by our method, corresponding to how well the primitive model can fit the true meshes.  ``NN Baseline'' is the nearest neighbor retrieval of shape in training set based on the embedded depth feature from our network. By enforcing rotation axis constraints (``3D-PRNN + rot loss''), our 3D-PRNN achieves better performance, which conforms with the learning curve as shown in Fig.~\ref{fig:learning}.  Though both nearest neighbor and 3D-PRNN are based on the trained encoding, 3D-PRNN outperforms NN Baseline for table and nightstand, likely because it is able to generate a greater diversity of shapes from limited training data. We compare with the voxel-based reconstruction of Wu et al.~\cite{wu20153d}, training and testing their method on the same data using publicly available code.  Since Wu et al. generate randomized results, we measure the average result over ten runs.  Our method performs similarly to Wu et al.~\cite{wu20153d} on the IoU measure.  Wu et al. performs better on surface distance, which is less sensitive to alignment but more sensitive to details in structures.  The performance of our ground truth primitives confirms that much of our reduced performance in surface distance is due to using a coarser abstraction (which though not preserving surface detail has other benefits, as discussed in introduction). 

\begin{table}
\begin{center}
\resizebox{0.85\linewidth}{!}{
\begin{tabular}{|c|c|c|c|}
\hline
 & chair & table & night stand\\
\hline\hline
GT prim & 0.473&0.533 & 0.657\\
\hline
NN Baseline&\textbf{0.269} & 0.220 & 0.256\\
Wu et al.~\cite{wu20153d} (mean) &0.253 &0.250 &\textbf{0.295} \\
3D-PRNN & 0.245 & 0.188 & 0.204\\
3D-PRNN + rot loss &  0.238 & 0.\textbf{263}& 0.266\\
\hline
\end{tabular}}
\end{center}
\vspace{-0.05in}
\caption{\label{iou-synthetic} Shape IoU evaluation in synthetic depth map in ModelNet. We explore two settings of 3D-PRNN with or without rotation axis constrains, and compare it with ground truth primitive and the nearest neighbor baseline. We also compare to the Wu et al.~\cite{wu20153d} deep network voxel generation method.}
\vspace{-0.1in}
\end{table}

\begin{table}
\begin{center}
\resizebox{0.85\linewidth}{!}{
\begin{tabular}{|c|c|c|c|}
\hline
 & chair & table & night stand\\
\hline\hline
GT prim & 0.049&0.044 & 0.044\\
\hline
NN baseline &0.075 & 0.089& 0.100\\
Wu et al.~\cite{wu20153d} (mean) &\textbf{0.045} & \textbf{0.035}&\textbf{0.057}\\
3D-PRNN & 0.074 & 0.080 & 0.104\\
3D-PRNN + rot loss &  0.074 & 0.078 & 0.092\\
\hline
\end{tabular}}
\end{center}
\vspace{-0.05in}
\caption{\label{surf-synthetic} Surface-to-surface distance evaluation in synthetic depth map in ModelNet. We explore two settings of 3D-PRNN with or without rotation axis constrains, and compare it with ground truth primitive and the nearest neighbor baseline.}
\vspace{-0.15in}
\end{table} 

\textbf{Real data (NYU Depth V2).} We also test our model on NYU Depth V2 dataset~\cite{silberman2012indoor} which is much harder than synthetic due to limited training data and the fact that depth images of objects are in lower resolution, noisy, and often occluded conditions. We employ the ground truth data labelled by Guo and Hoiem \cite{guo2013support}, where 30 models are manually selected to represent 6 categories of common furniture: chair, table, desk, bed, bookshelf and sofa. We fine-tune our network that was trained on synthetic data using the training set of NYU Depth V2. We report results on test set based on the same evaluation metric as the synthetic test shown in Table~\ref{iou-real} and \ref{surf-real}.  Since nightstand is less common in the training set and often occluded depth regions may be similar to those for tables, the network often predicts primitives in the shapes of tables or chairs for nightstands, resulting in worse performance for that class.  Sample qualitative results are shown in Fig.~\ref{qual:real}.

\textbf{3D Shape Segmentation.} Since our primitive based reconstructions are following meaningful part configurations naturally, another application where our method can apply is shape segmentation. Please refer to our supplemental material for shape segmentation task details and results, where we compare with state of the art methods as well. 

\begin{table}
\begin{center}
\resizebox{0.78\linewidth}{!}{
\begin{tabular}{|c|c|c|c|}
\hline
class & chair & table & night stand\\
\hline\hline
GT prim & 0.037&0.048 &0.020\\
\hline
NN baseline+ft & 0.118 & 0.176 & 0.162\\
NN baseline & \textbf{0.101} & \textbf{0.164} & \textbf{0.160}\\
3D-PRNN+ft & 0.112 & 0.168 & 0.192\\
3D-PRNN & 0.110 & 0.181 & 0.194\\
\hline
\end{tabular}}
\end{center}
\vspace{-0.05in}
\caption{\label{surf-real} Surface-to-surface distance evaluation in real depth map in NYUd v2. We explore two settings of 3D-PRNN with~(+ft) or without fine-tuning, and compare it with ground truth primitive and the nearest neighbor baseline.}
\vspace{-0.1in}
\end{table}

\begin{table}
\begin{center}
\resizebox{0.78\linewidth}{!}{
\begin{tabular}{|c|c|c|c|}
\hline
class & chair & table & night stand\\
\hline\hline
GT prim & 0.543& 0.435&0.892\\
\hline
NN baseline +ft & \textbf{0.171}  & \textbf{0.078}  & \textbf{0.286}\\
NN baseline & 0.145  & 0.076  & 0.262\\
3D-PRNN +ft & 0.158 & 0.075 & 0.081\\
3D-PRNN & 0.138 & 0.052 & 0.086\\
\hline
\end{tabular}}
\end{center}
\vspace{-0.05in}
\caption{\label{iou-real} Shape IoU evaluation in real depth map in NYUd v2. We explore two settings of 3D-PRNN with~(+ft) or without fine-tuning, and compare it with ground truth primitive and the nearest neighbor baseline.}
\vspace{-0.15in}
\end{table}

\textbf{Conclusions and Future Work.} We present 3D-PRNN, a generative recurrent neural network that uses recurring primitive based abstractions for shape synthesis. 3D-PRNN models complex shapes with a low parametric model, which advantages such as being capable of modeling shapes with fewer training examples available, and a large intra- and inter-class variance. Evaluations on synthetic and real depth map reconstruction tasks show that results comparable to higher degree of freedom representations can be achieved with our method. Future explorations include allowing various primitive configurations beyond cuboids (i.e. cylinders or spheres), encoding explicit joints and spatial relationship between primitives.

\section*{Acknowledgements}
This research is supported in part by NSF award 14-21521 and ONR MURI grant N00014-16-1-2007. We thank David Forsyth for insightful comments and discussion.

{\small
\bibliographystyle{ieee}
\bibliography{egbib}
}

\newpage
\clearpage
\appendix
\section{ $V_p$ in primitive fitting energy}
We define in Sec.~3.1 our primitive fitness energy as in Eq.~\ref{cost}. Where $V_p$ is the volumetric sampling ratio that is defined as the volume of the primitive $P$ over its number of sampled points $m$:

\begin{align}\label{V_P}
    V_p = \frac{s_xs_ys_z}{m}
\end{align}
The product $s_xs_ys_z$ is the volume of $P$. In our experiments, we set $m$ as a predetermined constant of $7\times7\times7$. 

\section{Derivatives of primitive fitting energy}

As we stated in Sec. 3.1, Eq.\ref{cost} is differentiable and can be solved using robust techniques (L-BFGS\cite{zhu1997algorithm}). In case $\exp\Big(-\frac{\|R(\theta) S p_m + T - q_n\|^2}{\sigma^2}\Big) < \xi$, derivatives are analytically defined. Otherwise, $E_p$ is a constant and the derivatives diminish.
The derivatives of Eq.\ref{cost} with respect to translation $T$, scale $S$, and rotation $\theta$ in the primitive parameter set $x$ are:
\begin{align}
    \frac{\partial E_p}{\partial T} =& -\sum_{m,n}V_p\exp\Big(-\frac{\|R(\theta) S p_m + T - q_n\|^2}{\sigma^2}\Big)\nonumber\\
    &\cdot \Big(-2\frac{R(\theta)Sp_m + T - q_n}{\sigma^2}\Big)
\end{align}
\begin{align}
    \frac{\partial E_p}{\partial S} =& -\sum_{m,n}V_p\exp\Big(-\frac{\|R(\theta) S p_m + T - q_n\|^2}{\sigma^2}\Big)\nonumber\\
    &\cdot \Big(-2\frac{R(\theta)Sp_m + T - q_n}{\sigma^2}\Big)\cdot R(\theta)p_m \nonumber\\
    &-\sum_{m,n}\frac{\partial V_p}{\partial S}\exp\Big(-\frac{\|R(\theta) S p_m + T - q_n\|^2}{\sigma^2}\Big)
\end{align}
where based on Eq.~\ref{V_P} we have:
\begin{align}
    \frac{\partial V_p}{\partial S} &= \Big[\frac{\partial V_p}{\partial s_x}, \frac{\partial V_p}{\partial s_y}, \frac{\partial V_p}{\partial s_z}\Big]\nonumber\\
    & = \Big[\frac{s_ys_z}{m}, \frac{s_xs_z}{m}, \frac{s_xs_y}{m}\Big]
\end{align}
\begin{align}
    \frac{\partial E_p}{\partial R(\theta)} =& -\sum_{m,n}V_p\exp\Big(-\frac{\|R(\theta) S p_m + T - q_n\|^2}{\sigma^2}\Big)\nonumber\\
    &\cdot \Big(-2\frac{R(\theta)Sp_m + T - q_n}{\sigma^2}\Big)\cdot Sp_m \frac{\partial R(\theta)}{\partial \theta}
\end{align}

$\theta$ is represented in Euler angles, thus $\frac{\partial R(\theta)}{\partial \theta}$ is the derivative of rotation matrix with respect to the rotation angles. Further details of the derivatives $\frac{\partial R(\theta)}{\partial \theta}$ can be found in~\cite{gallego2015compact}.

\section{Evaluation on primitive fitting}\label{eval}
We evaluate our primitive fitting method on the public semantic region dataset by Yi et al.~\cite{yi2016scalable}. The dataset contains detailed per-point labeling of model parts in ShapeNetCore~\cite{wu20153d} through an efficient semi-supervised labeling approach. We test our method on the test split of the chair category with 704 shapes, which contains part labeling of chair seat, back, arm and leg.

\begin{figure}\label{qual:seg}
\begin{center}
\includegraphics[width=0.97\linewidth]{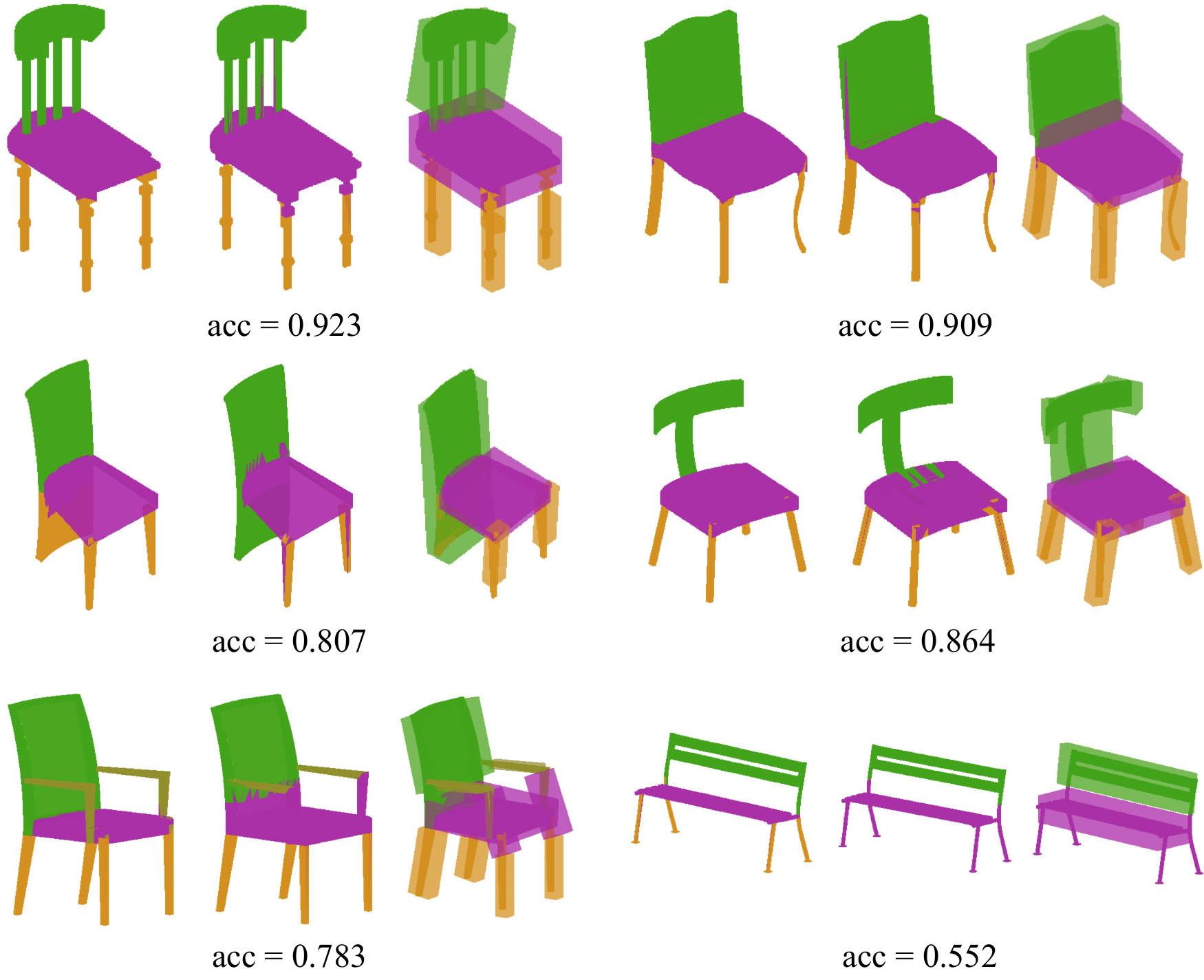}
\end{center}
   \caption{ Sample shape semantic labeling results on Yi et al's dataset~\cite{yi2016scalable}. For each row in each column, the left most shape represents the ground truth labeling of Yi et al's, the middle is the prediction by our method, and the right most one shows our primitive paring results being overlaid on the ground truth shape, the face labeling accuracy is also shown. The coloring indicates shape segments of chair back~(green), chair seat~(red), chair handle~(dark green) and chair leg~(yellow). The visualizations are based on face labeling. }
\end{figure}

We present our result based on the common metric of labeling accuracy at triangle face level, which measures the fraction of faces with correct labels. We project the ground truth per point labeling of shape parts into per face labeling through nearest neighbor search. Given our predicted primitive representation for the voxelized 3D shape, we randomly sample $N$ points (we set $N=10\times$ number of faces) on the shape, assign each point to the segment label of the nearest predicted primitive. We project our per-point segmentation to mesh faces based on a majority vote, i.e. if multiple points with different labels correspond to the same face, we label the face with the label suggested by the majority of the points. Since our primitive representation does not explicitly infer shape part label, we automatically re-label each part segmentation based on the majority vote of the ground truth labeling. We achieve the average face labeling accuracy of $\textbf{0.843}$. Qualitative results are shown in Fig.~\ref{qual:seg}. We observe that our primitive parsing method is able to decompose shapes into parts containing semantic meaning. Lower accuracy is often caused by 1) a single primitive that fits the shape but includes more than one type of the semantic meaning, see the shape in the second row, first column; 2) error prediction of regions with aggregation of faces, cylinder-shape handles of the bottom left shape contains more faces than the box-shape chair seat; 3) drawbacks of our method that fails to parse out slim shape segments, see the right bottom shape; 4) ground truth error results from projecting per-point labeling into per-face labeling.

\begin{figure*}
\begin{center}
\includegraphics[width=0.97\linewidth]{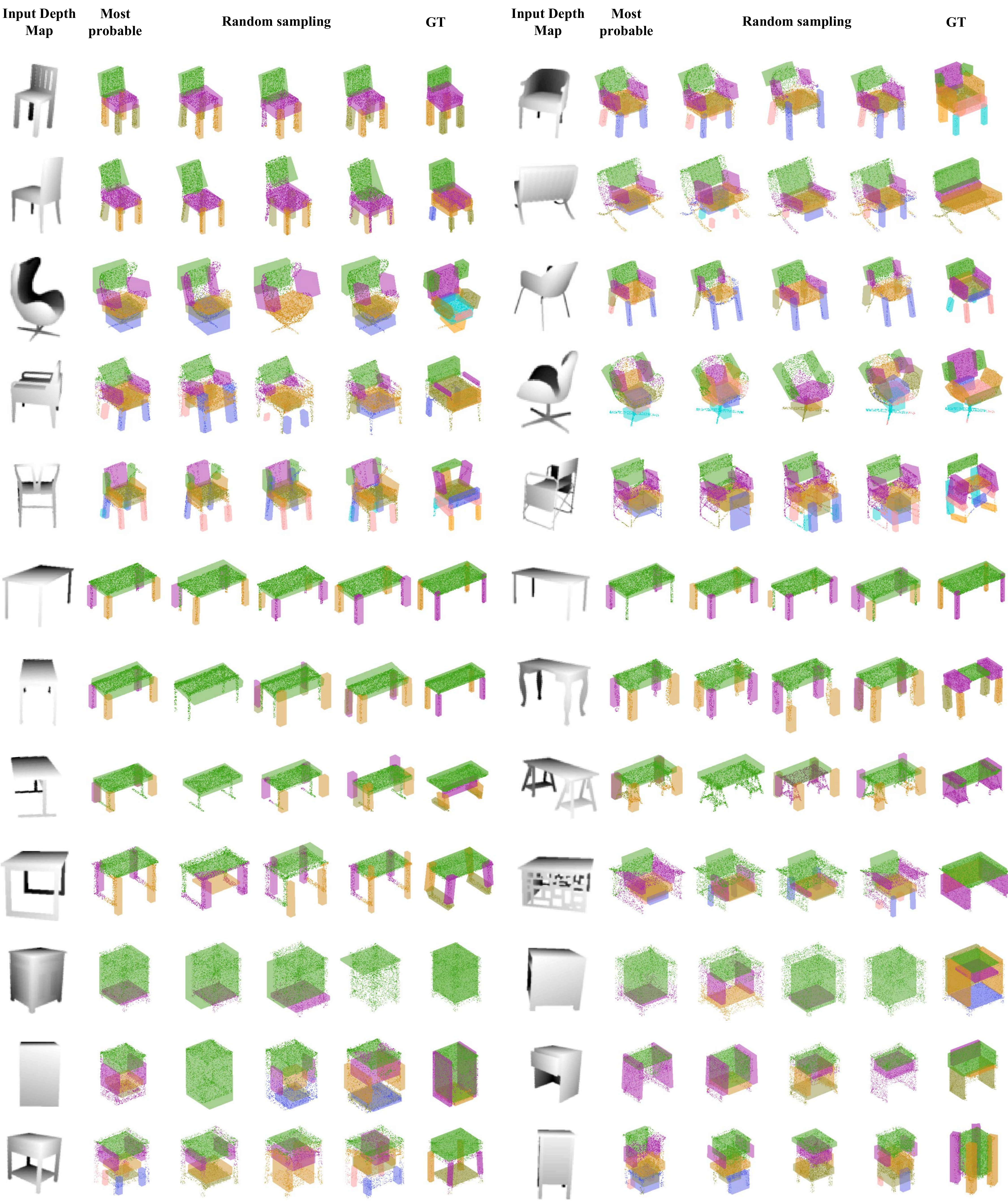}
\end{center}
   \caption{\label{qual:syn_supp} \textbf{Sample reconstruction from synthetic depth map in ModelNet}.We show the input depth map, with the most probable shape reconstruction from 3D-PRNN, and three successive random sampling results, compared with our groundtruth primitive representation. Each result is overlaid on the groundtruth cloud points.}

\end{figure*}

\begin{figure*}
\begin{center}
\includegraphics[width=0.97\linewidth]{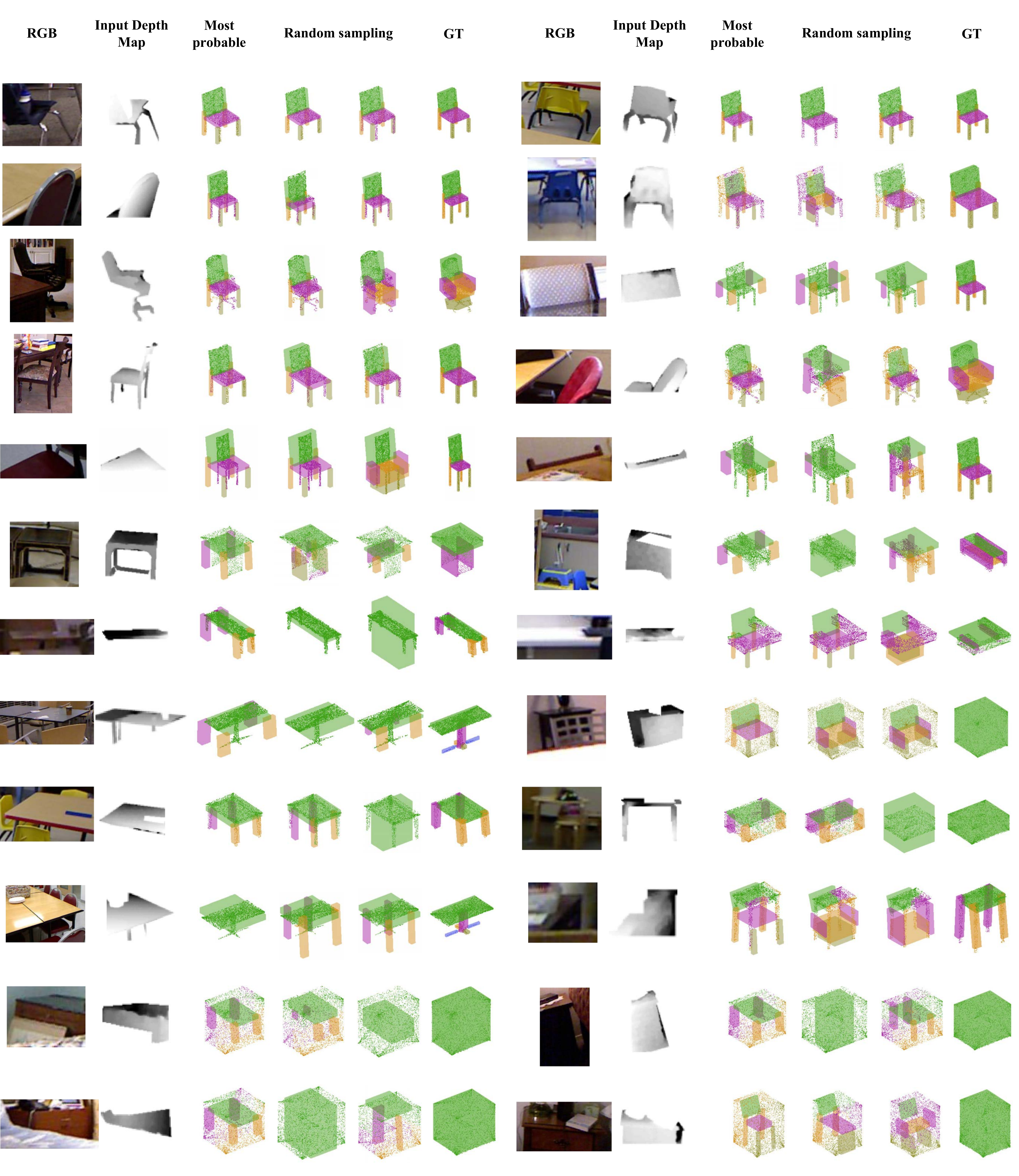}
\end{center}
   \caption{\label{qual:real_supp} \textbf{Sample reconstruction from real depth map in NYUdv2}.We show the input depth map, with the most probable shape reconstruction from 3D-PRNN, and two successive random sampling results, compared with our groundtruth primitive representation. Each result is overlaid on the groundtruth cloud points.}

\end{figure*}

\section{LSTMs sequential prediction order}
The recurrent generator of 3D-PRNN described in Sec.~4.1 has a pre-determined parameter set prediction ordering. At each time step we sample a single instance $x^s_i = [s_i, t_i, e_i]\in \mathbb{R}\times \mathbb{R}\times \{0,1\}$ drawn from the distribution $f(y_t)$. The sequence $x_t$ represents the parameters in the following order:
\begin{itemize}
\item  Time step 1, $x^s_x $ for the shape (width) and translation configuration on $x$ axis of the 1st primitive and the stopping sign.
\item Time step 2, $x^s_y $ for the shape (length) and translation configuration on $y$ axis of the 1st primitive and the stopping sign.
\item Time step 3, $x^s_z $ for the shape (height) and translation configuration on $z$ axis of the 1st primitive and the stopping sign.
\item Time step 4, $x^s_x $ for the shape (width) and translation configuration on $x$ axis of the 2nd primitive and the stopping sign.
\item Time step 5, $x^s_y $ for the shape (length) and translation configuration on $y$ axis of the 2nd primitive and the stopping sign.
\item $\ldots$ (sequential prediction of $x_t$)
\item Stop when "End of Generation" is predicted.
\end{itemize}

Note that the above sequence is for the primitive size parameters. We simultaneously predict rotation parameter $x^r$ and rotation axis $x^a$: at time step 1 we predict the rotation value $x^r_x$ on $x$ axis and a binary signal $x^a_x$ meaning whether there is rotation on $x$ axis or not for the first primitive, time step 2 predict $x^r_y$ and $x^a_y$ of the first primitive, then $x^r_z$ and $x^a_z$ of the first primitive. This simultaneous prediction also stops when "End of Generation" is predicted by the main LSTM prediction sequence outlined above. 


\section{Sampling procedure}
Note that an unexpected sample that is far from the mean of the distribution will cause accumulated error to the following predictions, during testing each time we sample from the first two most possible mixture component, in training we still perform random sampling on all mixture components. This strategy improves stability of our network performance in synthetic data case. In the real data case, we found that applying random sampling among all mixture components during both training and testing time can produce successive reasonable shapes. This is due to the fact that the ground truth shapes in real data are of simple structures that is easier to model by the network.

\section{Additional Results}
\subsection{Synthetic data}
Additional qualitative results of shape reconstruction from a single depth view for synthetic data are showed in Fig.~\ref{qual:syn_supp}.

\subsection{Real data}
Additional qualitative results of shape reconstruction from a single depth view for real data are showed in Fig.~\ref{qual:real_supp}.

\section{Application: shape segmentation}
Our primitive based reconstructions naturally align with semantic part configurations, and thus are directly applicable for shape segmentation tasks. To demonstrate this, we assume an input 3D shape is fully observed and use 3D-PRNN to reconstruct it as a collection of primitives. We then use the resulting primitives to semantcally segment the original input shape.

\textbf{Volumetric Encoder.}~The input 3D shape is represented as a $30\times30\times30$ binary voxel grid. We revise our previous depth image based encoder network to handle such voxelized input. Specifically, the first layer has $32$ kernels of size $7\times 7\times 7$ and stride $1$, with a LeakyRelu layer of $0.1$ slope in the negative part. The second layer consists of $64$ kernels of size $5\times 5\times 5$ (stride $1$), followed by the same setting of LeakyRelu and a max pooling layer of $2\times 2\times 2$ regions. The third layer has $128$ kernels of size $3\times 3$ (stride $1$) followed by similar LeakyRelu and max pooling layers. The next two fully-connected layers have $1024$ and $256$ neurons respectively. The output feature vector of dimension $1\times 256$ is then fed to the recurrent generator to predict a sequence of primitives. Note that the decoder and LSTM parts of the network remain the same. 

\textbf{Evaluation.}~We evaluate the performance of 3D-PRNN for the shape segmentation task on the COSEG dataset~\cite{wang2012active}. Since there is no groundtruth primitive representation of the dataset, for each shape, we automatically extract the tightest oriented box corresponding to each labeled segment and use it as a groudtruth primitive. Primitive ordering is pre-determined based on the height of each box center in a decreasing manner. Similar to the training scheme in the single depth reconstruction case, we first train our network on the random split of $70\%$ of the data, and validate it on $15\%$ of the data to choose the required number of training epochs. We then train on this $85\%$ of the data and perform tests on the remaining $15\%$ of the data which has never been seen by the network. 

Since the largest class of objects in COSEG is the chair category with 400 shapes, we test on this category. However, this is still a too small set for training an RNN. Hence, we first pre-train our network on ModelNet chair class with 889 shapes (with a $15\%$ validation split), then fine-tune our result on the COSEG chairs training set. This fine-tuning strategy increases our segmentation accuracy by $5\%$. We use ADAM to update network parameters with a learning rate of $0.001$, $\alpha = 0.95$, $\epsilon = e^{-6}$ and batch size $50$ for training and $20$ for fine-tuning.

\textbf{Results.} We present our result based on the metric same as Sec.~\ref{eval}. We compare the segmentation obtained on the most probable generation result of 3D-PRNN with the template-based segmentation result of Kim et al.~\cite{kim2013learning}, which is a state-of-the-art method that also fits oriented boxes to shapes for segmentation. We provide quantitative comparison in Table~\ref{tab-seg}. Note that our 3D-PRNN sometimes misses to predict some of the parts which numerically lowers our performance. Thus, we report both our overall performance and the average performance excluding such unpredicted parts. We also report the accuracy of the simple approach we used to generate the groundtruth primitive representations for training as this provides an upper bound for our method. In cases where 3D-PRNN predicts the correct number of primitives, it outperforms the method of Kim et al.~\cite{kim2013learning}. 

\begin{table}
\begin{center}
\resizebox{0.9\linewidth}{!}{
\begin{tabular}{|c||c|c|}
\hline
GT & Kim et al. & 3D-PRNN \\
\hline
0.896 & 0.829 &0.796\\
\hline
& \specialcell{Kim et al.\\(exc. unpredicted boxes)} & \specialcell{3D-PRNN\\(exc. unpredicted boxes)}\\
\hline
&0.836 & \textbf{0.859}\\
\hline
\end{tabular}}
\end{center}
\caption{\label{tab-seg} Shape segmentation result on the chair category of the COSEG dataset}
\end{table}

\end{document}